\pdfoutput=1

\documentclass[11pt]{article}

\usepackage[]{acl}

\usepackage{times}
\usepackage{latexsym}

\usepackage[T1]{fontenc}

\usepackage[utf8]{inputenc}

\usepackage{microtype}

\usepackage{graphicx}
\usepackage{comment}
\usepackage{multicol}
\usepackage{mdwlist}
\usepackage{multirow}
\usepackage{booktabs}
\usepackage{wrapfig}
\usepackage{amsmath,amssymb,bm}

\title{Unlearn What You Want to Forget: Efficient Unlearning for LLMs}

\author{Jiaao Chen \\
  Georgia Institute of Technology \\
  \texttt{jiaaochen@gatech.edu} \\\And
  Diyi Yang \\
  Stanford University \\
  \texttt{diyiy@cs.stanford.edu} \\}

\begin{document}
\maketitle
\begin{abstract}
Large language models (LLMs) have achieved significant progress from pre-training on and memorizing a wide range of textual data, however, this process might suffer from privacy issues and violations of data protection regulations. As a result, the ability to easily remove data related to individual users from such models while not deteriorating their predictive quality after the removal becomes increasingly important. To address these issues, in this work, we propose an efficient unlearning framework that could efficiently update LLMs without having to retrain the whole model after data removals, by introducing lightweight unlearning layers learned with a selective teacher-student objective into the transformers. In addition, we introduce a fusion mechanism to effectively combine different unlearning layers that learns to forget different sets of data to handle a sequence of forgetting operations. Experiments on classification and generation tasks demonstrate the effectiveness of our proposed methods compared to the state-of-the-art baselines\footnote{The codes are avaiable here: \url{https://github.com/SALT-NLP/Efficient_Unlearning/}}.

\end{abstract}

\begin{figure*}[t]
\centering
\includegraphics[width=2\columnwidth]{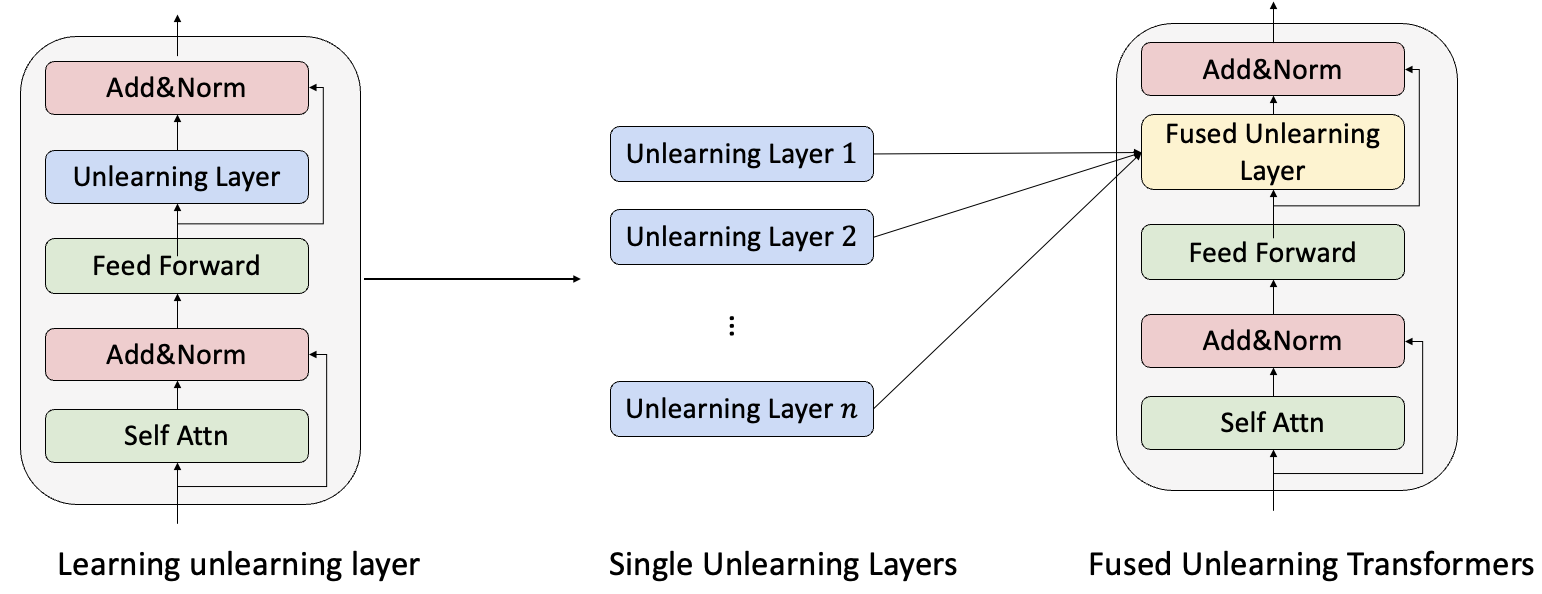}
\caption{Overall process of our EUL framework. The unlearning layers are plugged into transformer layers after the feed-forward networks. During training, only the unlearning layers are learned to forget requested data while the original LLMs remain unchanged. For every deletion request, an unlearning layer is learned first and then merged with other unlearning layers via our designed fusion mechanism to form the fused unlearning transformer which satisfies a series of deletion requests.
}
\label{Fig:model}
\end{figure*}

\section{Introduction}
Utilizing Large Language Models (LLMs) has become the dominant paradigm for various NLP applications \citep{brown2020language, palm,kojima2022large,ouyang2022training,brown2020language,Radford2019LanguageMA,lewkowycz2022solving,qin2023chatgpt,touvron2023llama} as LLMs memorize a vast amount of knowledge during pre-training or fine-tuning on a wide range of textual data  \citep{brown2020language,Radford2019LanguageMA,hoffmann2022training,webson2021prompt,min2022rethinking,liang2022holistic,carlini2022quantifying}. However, these data could contain sensitive information such as names, phone numbers, email addresses, and private clinical notes \citep{jang2022knowledge,kurmanji2023towards,kumar2022privacy}.
Extensive studies showed that LLMs could generate private information such as the Editor-in-Chief of MIT Technology Review including his family members, work address, and phone number \citep{carlini2022quantifying}. Recently, the EU's General Data Protection Regulation (GDPR) and US's California Consumer Privacy Act (CCPA) have also required the \textit{right to be forgotten}, introducing new regulations that require applications to support the deletion of user-generated content when requested by users \citep{sekhari2021remember,kumar2022privacy}. In light of this, it is essential to provide LLMs with an efficient and effective way to unlearn the information requested by users. 

Recent attention has been paid to the handling of such unlearning requests for LLMs through retraining and data pre-processing like SISA \citep{bourtoule2021machine,kumar2022privacy} where training data is stored in different isolated slices and each checkpoint is saved after training on each slice. When a deletion request is received, the respective data point will be removed from the slice, and the model checkpoint up to the data point will be used to further retrain the model. The effect of unlearning is often reflected by the model errors on the deleted data (models cannot predict the deleted data) \citep{kurmanji2023towards,jang2022knowledge}. Other works have also explored the design of algorithms that ensure differential privacy (DP) \citep{yu2021differentially,li2021large,anil2021large}. However, machine unlearning approaches like SISA \citep{bourtoule2021machine} usually require a significantly large amount of storage space \citep{bourtoule2021machine}, and DP methods could result in a slow convergence and significant deterioration in model performance \citep{nguyen2022survey}. In addition, both of them require retraining the whole model, which is extremely expensive and time-consuming considering the model scales of the current LLMs. These limitations also make them unable to dynamically deal with a sequence of unlearning requests which is often the need in real-world scenarios \citep{jang2022knowledge,nguyen2022survey}. 

To fill in these gaps, in this work, we propose an \textbf{E}fficient \textbf{U}nlearning method for \textbf{L}LMs (EUL) to efficiently unlearn what needs to be forgotten without completely retraining the whole model while retaining the performances of the models. Specifically, we propose a lightweight approach to learning the unlearning layer that is plugged into transformers through a selective teacher-student formulation \citep{kurmanji2023towards} within several updates,  without tuning the large language models. Additionally, we introduce a fusion mechanism to effectively combine the weights of different unlearning layers that learn to forget different sets of data to a single unified unlearning layer by minimizing a regression objective. This allows EUL to efficiently address a sequence of deletion operations. To demonstrate the effectiveness of our proposed EUL, we perform experiments on IMDB \citep{maas2011learning} and SAMSum \citep{gliwa2019samsum} in different settings compared to the state-of-the-art unlearning or model editing baselines. 
To summarize, our main contributions are threefold: 
\begin{itemize}\setlength{\itemsep}{1pt}
    \item We introduce an efficient unlearning method to remove the effect of required data in a lightweight way via a selective teacher-student formulation. 
    \item We design a fusion mechanism to merge unlearning layers that are learned to forget different sets of data into a single unlearning layer to deal with a sequence of removal operations.
    \item We conduct experiments on classification and generation tasks with backbone models of different scales in different settings, to illustrate the effectiveness of EUL.
\end{itemize}

\section{Related Work}

\subsection{Large Language Models}
Large language models have witnessed extensive progress recently \citep{brown2020language,Radford2019LanguageMA,smith2022using,rae2021scaling,chowdhery2022palm,touvron2023llama}, especially in terms of scaling up LLMs such as LLAMA \citep{touvron2023llama}, Megatron-turing NLG \citep{smith2022using}, Gopher~\citep{rae2021scaling}, and PaLM~\citet{chowdhery2022palm}. Other works have also achieved better performance with smaller models through longer training~\citep{hoffmann2022training}, instruction tuning \cite{wang2022self, zhou2023lima} and human feedback \citep{ouyang2022training}. However, recent studies have shown that training data, such as personally identifiable information like names, phone numbers, email addresses, and even bank account numbers \citep{carlini2021extracting,lee2021deduplicating,carlini2022quantifying,jagielski2022measuring}, can be easily extracted from LLMs because LLMs memorize the training data in billions of parameters \citep{carlini2022quantifying}. Our work is proposed to alleviate such issues by allowing efficient unlearning of the requested or private data from the learned parameters in LLMs.

\subsection{Machine Unlearning for Privacy}
To mitigate the privacy risks for LLMs, machine unlearning methods have been introduced to remove the contributions of training examples that users request to be erased by users \citep{bourtoule2021machine,chien2023efficient} including exact unlearning that retrains deep learning models on new datasets after removal \citep{bourtoule2021machine} and approximate unlearning \citep{izzo2021approximate,golatkar2020eternal,kurmanji2023towards,jang2022knowledge} which aims to modify the weights of trained models to produce a new set of weights that approximate the weights from retraining. The effect of unlearning is often reflected by the model errors on the deleted data (models cannot predict the deleted data) \citep{kurmanji2023towards,jang2022knowledge}. Another line of work has focused on Differential Privacy (DP) which ensures that user information in training data cannot be inferred \citep{dwork2008differential,yu2021differentially,li2021large,anil2021large,abadi2016deep}. However, both types of methods require retraining the whole model, which is extremely expensive and time-consuming, especially for large language models and even impacts the task performances \citep{anil2021large}. And thus they can not dynamically tackle sequences of deletion \citep{jang2022knowledge,nguyen2022survey}. To overcome these limitations, we introduce an efficient unlearning method as well as a fusion mechanism to \textbf{efficiently} and \textbf{dynamically} unlearn sequence of user data.

Our work is also related to model editing \citep{mitchell2021fast,belinkov2017neural,dai2021knowledge,wang2020k} while they usually focus on editing the model output based on several given linguistic structures or facts about the world instead of forgetting the required data.

\section{Efficient Unlearning for LLMs}
This section presents our designed \textbf{E}fficient \textbf{U}nlearning method for \textbf{L}LMs (EUL) which could efficiently and dynamically handle a sequence of deletion requests. The overall process is shown in Figure~\ref{Fig:model}. Formally, for a large language model $F(.)$ that is trained on a dataset $D=\{(x, y)\}$ where $x$ is textual data and $y$ is the corresponding label, and a deletion request to forget $D^f = \{(x^f, y^f\}$, our goal is to learn an updated model $F'(.)$ that satisfies the following \citep{kurmanji2023towards}:
\begin{equation}
\begin{aligned}
    I(F(D^f); F'(D^f)) &= 0 \\
    I(F(D^r); F'(D^r)) &= 1  
\end{aligned} \label{Eq:unlearning}
\end{equation}
where $D^r = D - D^f = \{(x^r, y^r)\}$ refers to the data we would like to retain, and $I(.)$ is the mutual information. Intuitively, we will update $F(.)$ with $F(.)$ to generate similar output for the data we want to retain while losing all information about making predictions on the data we want to forget.

\subsection{Learning to Forget via Unlearning Layers}
As the scales of current LLMs and the size of training data are usually large, updating all the parameters in the model $F(.)$ (e.g., re-training $F(.)$ on $D^r_i$) becomes extremely expensive. Inspired by recent advances in parameter-efficient fine-tuning \citep{houlsby2019parameter,chien2023efficient}, we model $F'(.)$ by $F(f(.))$ where $f(.; W)$ is an adapter with significant smaller amount of parameters $W$ compared to $F(.)$. And we would only update $f(.)$ to fulfill the unlearning requests.

To effectively achieve the unlearning goals in equation $\ref{Eq:unlearning}$, we minimize a selective teacher-student objective where the student model $F'(.) = F(f(.))$ is learned to follow the teacher model $F(.)$ on $D^r$ while disobeyed $F(.)$ on $D^f$:
\begin{equation}
\begin{aligned}
    L_{KL} = & \alpha \sum_{x^r} KL(F(x^r)||F(f(x^r))) \\ 
    &- \sum_{x_f} KL(F(x^f)||F(f(x^f)))
\end{aligned} \label{Eq:KLD}
\end{equation}
where $\alpha$ is a hyper-parameter to balance the trade-off between forgetting $x^f$ and retaining $x^r$. Intuitively, during training, $f(.)$ is leaned to minimize the KL-divergence between the output from the updated model and the original model on the data to retain while maximizing the KL-divergence between the output from them on the data to forget. 

To maintain the task performance, we optimize $f(.)$ for the task loss on the retain data:
\begin{equation}
\begin{aligned}
    L_{TASK} = \sum_{x^r} l(F(f(x^r)), y^r)
\end{aligned} \label{Eq:task}
\end{equation}
where $l(.)$ is the task-related loss, for example, cross-entropy loss, $-\log P(F(f(x^r)))$, for classification tasks. 

Furthermore, we also negate the original training objectives used in LLMs (e.g., masked language modeling objective \citep{t5}) to forget the knowledge related to the data, in order to forget in pre-trained parameters and ensure that the information in the forgotten data cannot be easily extracted from $F(.)$:
\begin{equation}
\begin{aligned}
    L_{LM} = - \sum_{x^f} l(F(f(x^f)))
\end{aligned} \label{Eq:lm}
\end{equation} 
where $l(.)$ is the language model loss used when pre-training $F(.)$, for example, masked language model loss, $-\log P(\hat{x}|x-\hat{x})$ ($\hat{x}$ are the randomly masked tokens). In our experiments, we utilize T5 models \citep{t5}. Thus we add an extra ``\textit{Predict the masked word}'' at the beginning of the input for this loss term.

Our final training objective is then the following:
\begin{equation}
\begin{aligned}
    L_{EUL} =  L_{KL} + \lambda L_{TASK} + \gamma L_{LM}
\end{aligned} \label{Eq:lm}
\end{equation}
where $\lambda$ and $\gamma$ are hyper-parameters. In practice, following \citet{kurmanji2023towards},  we alternate the updates for the data to be forgotten and the data to be retained to optimize \textit{min-max} terms in $L_{EUL}$ more stably. Specifically, we iteratively perform an epoch of updates on the data to be retained and then an epoch of updates on the data to be forgotten.

\subsection{Fusing Unlearning Layers}
To dynamically handle a sequence of unlearning requests and derive a unified model that could forget all of the requested data, we then introduce a fusion mechanism that could merge different unlearning layers $f_i(.; W_i)$ which are learned to forget $D^f_i = (X^f_i, Y^f_i)$ in the previous section into a single $f^m(.; W_m)$. Namely, we would like the output of $f^m(.)$ on $D^f_i$ being close to $f_i(.)$:
\begin{equation}
\begin{aligned}
    \min_{W_m} \sum_i || W_m^T X^f_i - W_i^T X^f_i||^2
\end{aligned} \label{Eq:lm}
\end{equation}
which is a linear regression problem and has a closed-form solution:
\begin{equation}
\begin{aligned}
    W_m = (\sum_i {X^f_i}^TX^f_i)^{-1}\sum_i({X^f_i}^TX^f_iW_i)
\end{aligned} \label{Eq:merge}
\end{equation}
Specifically, to derive the weights $W_m$ for the merged unlearning layer $f^m$, we would use the pre-computed inner product matrix of the hidden representations before the unlearning layers in LLMs of the forgotten data ${X^f_i}^TX^f_i$ and then compute $W_m$ following Equation~\ref{Eq:merge}. 

The fusion mechanism ensures efficiency and privacy as it could be performed without any extra training and only requires storing the inner product matrix of the representations of the data to be forgotten instead of the data itself.

\begin{table}[t]
\begin{center}
    \begin{tabular}{c|c|ccc}
    \toprule 
    \textbf{Dataset} &\textbf{Task} &\textbf{Train} &\textbf{Dev} &\textbf{Test} \\ \midrule \midrule
    IMDB & Classification & 20000 & 2000 &25000 \\ 
    SAMSum & Summarization & 14732 & 818 & 819 \\ \bottomrule
    \end{tabular}
\end{center} \caption{Dataset statistics for IMDB and SUMSum.} \label{Tab:dataset}
\end{table}

\begin{table*}[!t]
\centering
\small
\begin{tabular}{c|c|cccc|c}
\toprule
\textbf{Methods}             & \textbf{\# Forgot Data}                  & \textbf{Test Set}  $\uparrow$             & \textbf{Retained Set}   $\uparrow$                & \textbf{Forgot Set} $\downarrow$                & \textbf{MLM Loss} $\uparrow$     & \textbf{Time (s)} $\downarrow$    \\   \midrule \midrule               
\multicolumn{7}{c}{\textit{T5-base}} \\ \midrule \midrule
Original  & - &93.2 &100 &100 &1.46 &- \\ \midrule
Re-train &\multirow{6}{*}{0.5\%} & 92.8 & 100 & 92.5 &1.52 & 6685 \\
Fine-tune & &\textbf{93.0} &100 &96.5 & 1.47 & 4200 \\ 
SISA  & &92.4 &98.2 &91.5 &1.54  & 1580 \\ 
Reverse-Gradient & &92.0 &97.3 &68.6 & 1.56 & 4400  \\ 
MEND & & 92.2 &98.5 &73.5 &1.60 & \textbf{34} \\  
EUL$\dag$ & &\textbf{93.0} & \textbf{100} & \textbf{65.7} & \textbf{1.78} & 1200  \\  \midrule \midrule

Re-train & \multirow{6}{*}{1\%} &92.7 & 100 & 91.6 & 1.55  & 6610   \\
Fine-tune & &92.8 & 100 & 96.2 & 1.48 & 3950    \\ 
SISA  &  & 92.2 & 98.1 & 90.4 & 1.55  & 2930  \\ 
Reverse-Gradient & &91.5 & 96.4 & 67.4 &1.59 & 4166   \\ 
MEND &  & 91.3 & 95.5 & 74.6 & 1.62 & \textbf{62} \\
EUL$\dag$ & &\textbf{93.0} & \textbf{100}  & \textbf{64.4}  &\textbf{1.84} & 1526 \\ \midrule \midrule

Re-train &  \multirow{6}{*}{10\%} & 92.1 & \textbf{100} & 90.2 & 1.56 & 6026  \\
Fine-tune & & 92.0 & 100 & 95.8  & 1.52 & 3133 \\ 
SISA  & & 91.6 & 98.2 & 88.4 & 1.55 & 2010  \\ 
Reverse-Gradient & &91.0 & 96.5 & 65.4 & 1.62 & 3228  \\ 
MEND &  & 90.8 & 94.8 & 76.2 & 1.66 & \textbf{328} \\ 
EUL$\dag$ & & \textbf{92.2} & 99.0 & \textbf{57.2} & \textbf{2.01} & 1828  \\  \midrule \midrule 

\multicolumn{7}{c}{\textit{T5-3b}} \\ \midrule \midrule
Original &- & 97.0 & 100 & 100 &1.28 &-  \\ \midrule
Re-train & \multirow{6}{*}{0.5\%} & 96.6 & 100 & 94.8 & 1.30 & 26855 \\
Fine-tune & & \textbf{96.7} & 100 & 96.2 & 1.28 & 20465 \\ 
SISA  & &95.0 & 97.2 & 94.1 & 1.33 & 16503  \\ 
Reverse-Gradient & &93.3 &96.5 & 78.9 & 1.42 & 21826  \\ 
MEND & & 93.0 &95.8 &89.5 &1.30 &\textbf{4980} \\ 
EUL$\dag$ & &96.5 & \textbf{100} & \textbf{70.2} &\textbf{1.66} & 9240 \\  \midrule \midrule

Re-train & \multirow{6}{*}{1\%} &96.3 &100 &94.2 & 1.30 & 25280 \\
Fine-tune & &\textbf{96.5} &100 &96.0  &1.28 & 18466  \\ 
SISA  &  & 93.8 & 96.8 & 92.7 &1.35 &15680 \\ 
Reverse-Gradient & &92.5 & 96.0 & 80.1 &1.46 & 18800  \\ 
MEND & & 92.8 &95.0 &84.4 &1.48 &\textbf{6600}   \\
EUL$\dag$ & &\textbf{96.5} &\textbf{100} &\textbf{67.5} &\textbf{1.72} & 9840  \\  \midrule \midrule

Re-train & \multirow{6}{*}{10\%} &96.0 & 100 &93.5 &1.31 &22140  \\
Fine-tune & & \textbf{96.2} &100 &94.0 & 1.30 & 16752 \\ 
SISA  & & 93.0 & 95.5 & 92.2 & 1.35 & 14180 \\ 
Reverse-Gradient & &91.9 & 95.2 &68.4 & 1.46 & 17850  \\ 
MEND & & 92.0 &94.2 &78.5 &1.50 & 12072   \\ 
EUL$\dag$ & &96.0 &\textbf{100} & \textbf{60.8} &\textbf{1.92} &\textbf{10460}  \\  \bottomrule    
\end{tabular} \caption{Performances on IMDB for T5-base and T5-3B after unlearnling different 
 number of privacy-related data. $\dag$ refers to our model. All the results are averaged over 5 random runs.
} \label{Tab:imdb}
\end{table*}

\section{Experiments}

\subsection{Datasets}
We conduct experiments on both classification and generation tasks. For the classification task, we utilize the IMDB dataset\citep{maas2011learning}, which is a sentiment classification dataset consisting of users' reviews of movies, directors, actors, etc. For the generation task, we use SAMSum \citep{gliwa2019samsum}, which is a recent popular conversation summarization dataset consisting of conversations between different speakers. The dataset statistics are shown in Table~\ref{Tab:dataset}.

We choose these two datasets because they are widely used \citep{wang2021entailment,yang2019xlnet,qin2023chatgpt,ji2023survey,wei2021finetuned,sanh2021multitask,chen2022humanintheloop} to evaluate large language models and both datasets are related to cases where the user might require to remove their data, for example, removing all the reviews of a specific movie or removing all the conversations from one specific speaker. 

In experiments, we use the pre-trained NER models from AllenNLP\footnote{\url{https://demo.allennlp.org/}} to extract all the entities (names) in IMDB and directly use the speakers' names in SAMSum and simulate the unlearning requests to remove all the data from or related to certain names. Moreover, we substitute all the names in the dev and test set with special tokens.

\begin{table*}[!t]
\begin{center}
\small
\begin{tabular}{c|c|cccc|c}
\toprule
\textbf{Methods}             & \textbf{\# Forgot Data}                  & \textbf{Test Set}  $\uparrow$             & \textbf{Retained Set}   $\uparrow$                & \textbf{Forgot Set} $\downarrow$                & \textbf{MLM Loss} $\uparrow$     & \textbf{Time (s)} $\downarrow$    \\   \midrule \midrule     
\multicolumn{7}{c}{\textit{T5-base}} \\ \midrule \midrule
Original  & - &47.2/23.5/39.6  & 71.4/42.6/62.7  & 70.2/42.2/62.7  & 1.37 &- \\ \midrule
Re-train & \multirow{5}{*}{0.5\%}  & 46.8/23.0/38.1  & 71.7/42.8/62.4     & 42.4/23.2/42.0  & 1.40 & 28000\\
Fine-tune & & 46.6/23.2/38.1 & \textbf{72.5/44.7/65.2} & 58.8/34.1/54.1 & 1.38 & 27120\\ 
SISA  & &44.2/22.0/37.4 & 70.5/41.6/60.5 & 41.4/23.0/40.8   & 1.48  & 22582  \\ 
Reverse-Gradient & & 43.2/20.9/35.8  & 68.8/40.2/58.5 & 42.3/21.4/38.1  & 1.64 &  28800 \\ 
EUL$\dag$ &  & \textbf{46.8/23.0/38.5}  & 71.5/42.4/63.3  & \textbf{38.4/20.2/37.2} & \textbf{1.88} & \textbf{17060}  \\  \midrule \midrule

Re-train &\multirow{5}{*}{1\%}  & 45.4/22.8/37.5  & 72.4/43.0/62.8     & 42.2/22.8/41.6  & 1.44 & 26855  \\
Fine-tune & & \textbf{46.4/23.2/38.1} & \textbf{72.9/43.6/64.0} & 56.4/31.8/52.7 & 1.40 & 27210\\ 
SISA  & &43.1/21.1/36.8 & 69.8/40.2/60.0 & 41.4/23.0/40.8   & 1.50  & 22420  \\ 
Reverse-Gradient & & 42.0/20.0/34.6  & 68.8/40.2/58.5 & 42.3/21.4/38.1  & 1.64 &  27700 \\ 
EUL$\dag$&  & 46.5/22.8/38.0  & 71.5/42.4/63.3  & \textbf{35.8/19.0/36.2} & \textbf{1.95} & \textbf{16820}  \\  \midrule \midrule

Re-train & \multirow{5}{*}{10\%} & 44.2/21.2/35.8  & 70.4/41.2/60.5  & 41.4/21.4/40.0  & 1.48 & 26155  \\
Fine-tune & & 45.2/22.1/36.6 & \textbf{71.1/42.6/62.9} & 51.5/28.6/50.0 & 1.43 & 27510 \\ 
SISA  & &41.8/19.6/33.8 & 68.3/38.8/58.8 & 40.2/20.1/38.9   & 1.55  & 20790  \\ 
Reverse-Gradient & & 40.8/18.4/33.0  & 66.6/38.3/55.5 & 38.0/19.4/36.6  & 1.71 &  27240 \\ 
EUL$\dag$ &  & \textbf{45.8/22.4/37.8}  &70.9/42.0/62.3  & \textbf{33.0/18.3/33.0} & \textbf{2.23} & \textbf{15000}  \\  \midrule \midrule

\multicolumn{7}{c}{\textit{T5-3b}} \\ \midrule \midrule
Original  & - &53.6/29.6/45.1  & 78.5/47.6/66.1  & 74.2/43.5/64.9  & 1.30 &- \\ \midrule
Re-train & \multirow{5}{*}{0.5\%}  & 52.8/28.8/44.0  & 77.4/46.1/65.4     & 50.4/27.2/43.0  & 1.34 & 84480\\
Fine-tune & & 53.3/29.0/44.4 & \textbf{78.0/47.1/65.8} & 60.2/36.1/55.7 & 1.30 & 83600\\ 
SISA  & &51.7/27.2/40.8 & 74.8/44.8/63.5 & 49.4/26.8/42.2    & 1.33  & 75000  \\ 
Reverse-Gradient & & 50.6/25.9/39.9  & 72.8/42.0/62.8 & 44.3/23.1/39.0  & 1.44 &  83200 \\ 
EUL$\dag$ &  & \textbf{53.6/29.4/44.8}  & 77.5/46.3/66.6  & \textbf{41.0/21.8/38.2} & \textbf{1.67} & \textbf{60430}  \\  \midrule \midrule

Re-train &\multirow{5}{*}{1\%}   & 52.0/28.2/42.8  & \textbf{76.7/45.8/64.8}     & 49.6/26.6/42.1  & 1.35 & 82440\\
Fine-tune & & 52.5/28.5/43.6 & 76.2/45.5/64.2 & 56.8/32.2/52.4 & 1.32 & 81135 \\ 
SISA   & &50.0/26.1/38.9 & 72.3/43.1/61.1 & 49.0/25.8/41.1    & 1.38  & 73550  \\ 
Reverse-Gradient & & 48.6/24.3/37.2  & 70.6/41.5/60.9 & 42.2/22.0/37.7  &1.45 &  82485 \\ 
EUL$\dag$  &  & \textbf{53.3/29.0/44.4}  & 76.4/45.3/64.3  & \textbf{38.4/19.9/36.0} & \textbf{1.74} & \textbf{60880}  \\  \midrule \midrule

Re-train & \multirow{5}{*}{10\%} & 50.8/26.4/40.5  & 74.2/45.0/63.2     & 48.2/25.5/41.4  & 1.38 & 81010\\
Fine-tune & & 51.4/27.2/41.9 &\textbf{75.2/45.3/64.0} & 52.1/29.8/49.9 & 1.35 & 81800 \\ 
SISA  & &48.2/24.5/36.0 & 70.4/40.5/59.6 & 41.2/23.5/40.0    & 1.40  & 70400  \\ 
Reverse-Gradient & & 44.7/22.0/34.2  & 68.5/40.9/58.8 & 40.9/21.0/36.5  &1.49 &  82070 \\ 
EUL$\dag$ &  & \textbf{52.0/28.4/42.6}  & 74.9/45.0/63.6  & \textbf{36.2/18.6/34.7} & \textbf{1.78} & \textbf{59900}  \\  \bottomrule
\end{tabular}  
\end{center} \caption{Performances on SAMSum for T5-base and T5-3B after unlearnling different 
 number of privacy-related data.  $\dag$ refers to our model. All the results are averaged over 3 random runs. The performance on Test, Retained and Forgot Set are ROUGE-1/2/L scores.} \label{Tab:samsum}
\end{table*}
\subsection{Evaluation Metrics}
To evaluate the performances, following \citet{kurmanji2023towards}, we measure several metrics: (1) \textbf{Performance on the test set}: The task-related performance on the test set, namely, accuracy for IMDB and ROUGE for SAMSum. This measures whether the unlearning algorithms affect the model performance or not.  (2) \textbf{Performance on the retained set}: The task-related performance on the data to be retained. This measures whether the unlearning algorithms forget the data that need to be retained. Higher performance means that the model remembers the data that is not to be forgotten. (3) \textbf{Performance on the forgot set}: The task-related performance on the data to be forgotten. This measures whether the unlearning algorithms effectively forget the data requested to be forgotten. Lower performance means that the model is better at forgetting the data. (4) \textbf{MLM Loss}: The masked language model loses on the data to be forgotten where related entities or actions are masked. This is achieved by adding ``\textit{Predict the masked word}'' in the beginning. This measure whether the information in the data that needs to be forgotten can be extracted from the LLMs. 
Higher MLM loss means that it is harder to extract such information from the models. (5) \textbf{Updating time}: The time to update the original model in the forgetting process. 

\begin{table*}[t]
\begin{center}
\begin{tabular}{c|ccc|c}
\toprule
\textbf{Methods}                           & \textbf{Test Set}  $\uparrow$             & \textbf{Retained Set}   $\uparrow$                & \textbf{Forgot Set} $\downarrow$                & \textbf{Updating Time (s)} $\downarrow$    \\   \midrule \midrule             
Original   &91.8 &100 &91.2 &- \\ \midrule
Re-train & 92.5  & \textbf{100} & 12.6  & 6026  \\
Fine-tune  & 92.3 & 100 & 26.8  & 3133 \\ 
SISA   & 92.2 & 98.2 & 12.6  & 1510  \\ 
Reverse-Gradient  &92.8 & 98.6 & 9.0  & 3228  \\ 
MEND   & 92.2 & 97.8 & 16.8  & \textbf{328} \\ 
EUL$\dag$  & \textbf{93.0} & 99.0 & \textbf{5.0} & 1828  \\   \bottomrule    
\end{tabular} 
\end{center} \caption{Performances on IMDB for T5-base after unlearnling 10\% wrong-labeled data.  $\dag$ refers to our model. All the results are averaged over 5 random runs.} \label{Tab:imdb_error}
\end{table*}

\begin{figure*}[t]
\centering
\includegraphics[width=2.0\columnwidth, height = 0.32\textwidth]{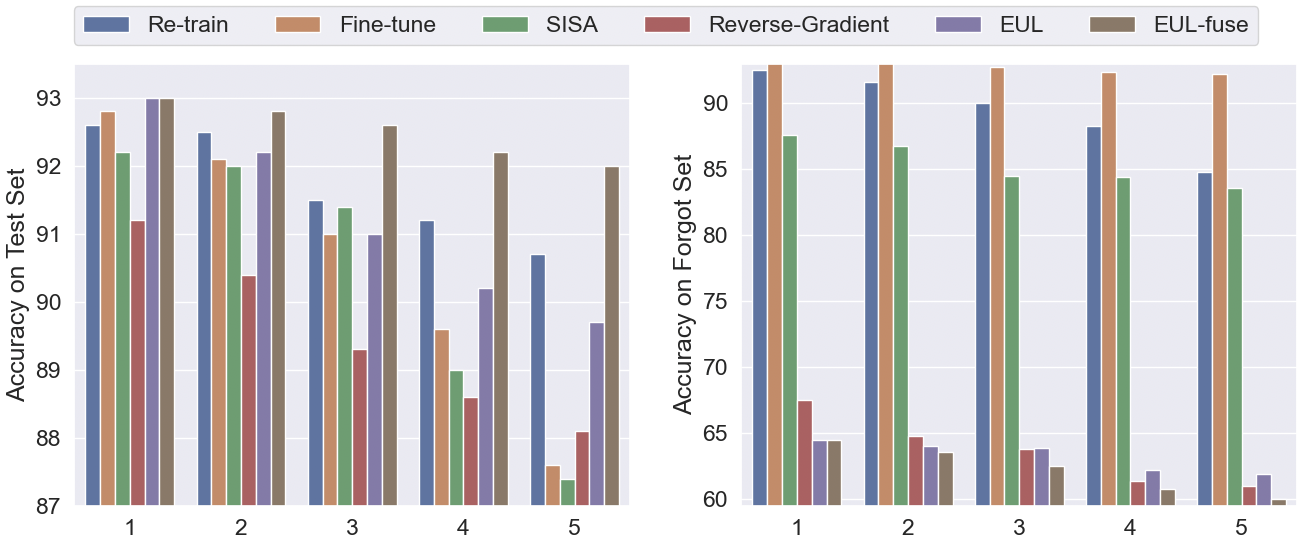}
\caption{Sequentially unlearnling 1,2,3,4,5 different sets of data for T5-base on IMDB. The results are accuracy on the test set and the accuracy on the forgot set averaging across different orderings. Every single set contains 1\% of the training data.}
\label{Fig:imdb_sequence}
\end{figure*}

\subsection{Baselines}
We compare our EUL with several baseline methods: \textbf{Re-train} \citep{kumar2022privacy}: Re-training the model from scratch on the data to be retained without any forgotten data.  \textbf{Fine-tune} \citep{kurmanji2023towards}: Fine-tuning the original model on the data to be retained without any forgotten data. \textbf{SISA} \citep{kumar2022privacy}: Sharded, Isolated, Sliced, and Aggregated training where multiple models are trained independently on disjoined shards, and its slices and model checkpoints are saved for each slice.  When forgetting certain data, the corresponding data point is deleted from its slice, and the model checkpoint up to the data point is used to further retrain the model. \textbf{Reverse-Gradient} \citep{liu2022continual}: Fine-tuning the original model on both retained data and forgot data while negating the gradient for the forgot data.  \textbf{MEND} \citep{mitchell2021fast}: Editing the model to generate output following the given examples. To adapt the model in the unlearning setting, we reverse the labels for data in classification tasks as input to MEND. However, it is infeasible to apply MEND to summarization tasks as it is hard to design the new output to perform the editing.

\subsection{Model Settings}
For all the experiments, we use T5 models (T5-base and T5-3b) \citep{t5} as the backbone models. For SISA, we follow \citet{kumar2022privacy} to split the dataset. For our unlearning layers, we only tune 0.5\% \citep{chen2023parameter} of the parameters. The $\alpha = 0.8$, $\lambda = 1.0$ and $\gamma = 0.2$ are selected from grid searching $\{0.1, 0.2, 0.5, 0.8, 1.0\}$. We set the linear decay scheduler with a warmup ratio of 0.06 for training. The maximum sequence length is 128 for IMDB and 800 for SAMSum. The batch size was 256 for base models and 128 for 3b models on IMDB and 8 for base models and 2 for 3b models on SAMSum. The maximum learning rate was $5e-5$ and the maximum number of training epochs was set to be $3$ or $5$. All the experiments were performed using 8 A100 GPUs.

\begin{table}[t]
\begin{center}
\small
\begin{tabular}{c|c|ccc}
\toprule
\textbf{Metric}                           & \textbf{EUL} \          & \textbf{-KL} \              & \textbf{-TASK} \  & \textbf{-LM}\           \\   \midrule \midrule 
Test Set  $\uparrow$     &\textbf{93.0} &91.4 &91.0 &92.4  \\ 
Retained Set  $\uparrow$   & \textbf{100} &100 & 97.4&99.0\\ 
Forgot Set $\downarrow$   & \textbf{65.7} & 90.8&  67.4 &69.0  \\  
MLM Loss $\uparrow$      &\textbf{1.78}   & 1.75 & 1.78 & 1.50  \\     \bottomrule  
\end{tabular} 
\end{center} \caption{Performances on IMDB for T5-base after removing 0.5\% privacy-related data. We remove one objective at a time from our EUL methods. } \label{Tab:ablation}
\end{table}

\begin{table}[t]
\begin{center}
\small
\begin{tabular}{c|ccc}
\toprule
\textbf{Models}                           & \textbf{Set 2} \          & \textbf{Set 2, 1}   & \textbf{Set 2, 1,3}               \\   \midrule \midrule 
Re-train & 92.7/91.4	&92.5/90.8	&91.3/90.0 \\
Fine-tune & 92.8/96.0	&92.1/94.0	&91.0/93.3\\
SISA &92.2/90.4	&92.0/87.8	&91.2/85.8 \\
Reverse-Gradient &91.5/67.9	& 90.5/67.2	& 89.8/66.0 \\ \midrule
EUL &\textbf{93.0/64.6}	& 92.1/64.8	& 91.0/64.2 \\
EUL-fuse &\textbf{93.0/64.6}	&\textbf{92.8/62.2}	& \textbf{92.4/60.8}\\
\bottomrule  
\end{tabular} 
\end{center} \caption{Accuracy on the test/retained set of after unlearning sets of data following a sequence (set 2 -> set 1 -> set 3). } \label{Tab:sequence}
\end{table}

\subsection{Results}

\paragraph{Unlearning Privacy-related Data on IMDB} We request the T5-base and T5-3b models that are fine-tuned on the IMDB dataset to unlearn 0.5\%, 1\% and 10\% of the training data. The data to be forgotten is randomly selected based on the names of movies, actors, actresses, directors, etc. For example, the model might need to forget all the data points related to ``\textit{Lena Numan}''.  This simulates the cases where people/companies request to remove all the data related to them. The performances are displayed in Table~\ref{Tab:imdb}. 

After unlearning the requested data from T5-base models, the re-training method hurts the accuracy (e.g., a 1.1 accuracy drop when forgetting 10\% data) on the test set because there is fewer data for training, and the accuracy on the retained set keeps unchanged (100\%) probably because the model memorizes the retained data. The accuracy on the forgot set drops after re-training (e.g., 92.5 compared to 100 when unlearning 0.5\% of the data), showing that the model is forgetting the requested data, and the masked language model loss increases (e.g., increasing 0.06 when unlearning 0.5\% of the data), indicating that it is harder to extract the information of the forgot data after re-training. The fine-tuning method shows better test accuracy with less updating time, however, it is worse in terms of forgetting the data. Even though SISA takes significantly less time (only costing around 1/3 of the time compared to re-training)  to derive the updated model that forgets the requested data, it receives lower accuracy on the test and retained set, which means that the model prediction abilities get worse because of failing to remember the retained data. When reversing the gradients for the data to be forgotten, the updated model gets better at forgetting with lower test accuracy. The model editing method, MEND, shows better overall performance on nearly all the metrics but it requires extra data to train a model editing module to edit the original model, making the method hard to be generalized to new models and settings. Our EUL approach boosts all the metrics with faster speed to update the model compared to previous unlearning baselines after removing different numbers of privacy-related data (e.g., achieving the lowest accuracy (65.6\%) on forgot set while keeping the best test accuracy (93.0\%) and 100\% retained accuracy with 1/6 of the updating time compared to re-training when forgetting 0.5\% of the data), suggesting that our designed unlearning layers that are learned with tailored objectives could efficiently update the LLMs to forget the required data and remain the abilities to perform the tasks. When the size of the backbone model scales up to 3b, the improvements of our EUL are consistent, indicating that our methods could still forget what the user requests even for larger models that are better at memorizing data.

\paragraph{Unlearning Privacy-related Data on SAMSum} We unlearn 0.5\%, 1\% and 10\%  training data from T5-base and T5-3B models that are fine-tuned on the SAMSum dataset. The data to be forgotten is randomly selected based on the speaker names. For example, the model might need to forget all the conversations from ``\textit{Jack}''. This simulates the cases where people request to remove all the data generated by them. The performances are shown in Table~\ref{Tab:samsum}. Similarly, our EUL method consistently achieves the best overall performances by effectively forgetting the requested data while remembering the retained data and keeping the test ROUGE scores with significantly less amount of training time. This indicates that our objectives could also be generalized to generation tasks.

\paragraph{Unlearning Mislabeled Data on IMDB} We also test a setting where the data to be forgotten is those with wrong labels. In experiments, we randomly change the labels for 10\% of the training data and then request the model to unlearn their impact. This simulates the cases where we improve the models that are trained on noisy data by unlearning the mislabeled data \citep{kumar2022privacy}. We report the performances with T5-base models in Table~\ref{Tab:imdb_error}. We observe that the accuracy of the test set of the original model is affected by the mislabeled data. And our EUL is the most effective approach to unlearn and remove the negative impact of those mislabeled data to achieve the best test accuracy.

\paragraph{Sequence of Removals} We test baseline and our methods in a setting where a sequence of unlearn requests are received, i.e., the models need to forget different sets of data sequentially. In experiments, we sequentially unlearn 1,2,3,4,5 sets of data from T5-base model on IMDB dataset. For every unlearn length, we test with all the possible sequences and average the accuracy on the test set and the forgot set. For example, when the length of the forgetting requests are 2 (set 1, set 2), we test on the sequence (set 1 -> set 2) and sequence (set 2-> set 1) and average the final performances. We show the results (accuracy on the test/retained set) of one possible sequence whose length is 3 (set 2 -> set 1 -> set 3) in Table~\ref{Tab:sequence} as an example. Averaged performances over different sequence lengths are visualized in Figure~\ref{Fig:imdb_sequence}. 
EUL means that we keep one unlearning layer to sequentially unlearn different sets of data and EUL-fuse means that for every set of forgot data we learn separate unlearning layers and then merge them into a single unlearning layer via our proposed fusion mechanism. The results demonstrate that our proposed fusion method that combines different unlearning layers could effectively handle the sequence of deletion (achieving higher accuracy on the test set and lower accuracy on the forgot set.) especially when the sequence length gets longer compared to baseline models. 


\begin{table}[t]
\begin{center}
\begin{tabular}{c|cc}
\toprule
\textbf{Models}                           & \textbf{IMDB} \          & \textbf{SAMSum}                \\   \midrule \midrule 
Original & \textbf{0.542} & \textbf{0.510} \\ \midrule 
Re-train & 0.550 & 0.522 \\
Fine-tune & 0.568 & 0.525 \\
SISA & 0.585 & 0.530 \\
Reverse-Gradient &0.626 &0.588 \\
EUL &0.566 & 0.530 \\
\bottomrule  
\end{tabular} 
\end{center} \caption{Accuracy from a trained binary classifier to predict whether an input data belongs to the retained set or the forgot set.} \label{Tab:mia}
\end{table}

\subsection{Ablation Studies}

\paragraph{Removal of Objectives}
We perform ablation studies to show the effectiveness of each designed objective in EUL by removing each of them when learning the unlearning layers in Table~\ref{Tab:ablation}. Compared to EUL which utilizes all of the learning objectives, removing each of them would result in a performance drop, which demonstrates every component contributes to the final performance. Specifically, removing $L_{KL}$ would increase the accuracy of the forgot set, indicating that $L_{KL}$ is the main factor to forget the requested data. Removing $L_{TASK}$ from EUL would drop the accuracy on the test set, suggesting that $L_{TASK}$ is essential to maintain task performance. Removing $L_{LM}$ decreases the MLM Loss, showing that $L_{LM}$ is the main objective to avoid the extraction of the requested information.

\paragraph{Member Inference Attack}
We further perform Member Inference Attack (MIA) \citep{kurmanji2023towards} on IMDB and SAMSum when unlearn 1\% privacy-related data for T5-base models. Specifically, we test the accuracy of a binary classifier which is trained to predict whether the input data belong to the forgotten set or the retained set based on their representations after the final layer of the T5 model. An accuracy closer to 0.5 means that it is hard for the classifier to predict the groups of the input data. The accuracies are shown in Table~\ref{Tab:mia}. We found that the classifiers could not converge so well on the training set and always had a low accuracy on the test set both before and after unlearning (e.g., 0.542 before unlearning and 0.566 after our EUL unlearning on IMDB). These showed that the randomly deleted data could not be easily inferred both before and after our EUL unlearning.

\section{Conclusion}
In this work, we propose EUL, an efficient unlearning method for LLMs that could efficiently and effectively unlearn the user-requested data via learning unlearning layers through the selective teacher-student objective. We further introduce a fusion mechanism that could merge different unlearning layers into one unified layer to dynamically unlearn a sequence of data. Experiments on different settings (different datasets, different model sizes, different forget set sizes) demonstrated the effectiveness of our proposed EUL method compared to state-of-the-art baselines.

\section{Limitations}
In this work, we mainly perform experiments on T5-base/3b models with fine-tuned tasks. We encourage future work to explore how to update different backbone models with larger sizes such as LLAMA models or even close-sourced models like ChatGPT to forget the requested data such asemn privacy-related data, toxic data, or misinformation in the pre-training corpus. Also, we mainly follow the previous work to measure the unlearning through performance on the test set, retained set, and forgot set, together with the MLM loss. Future work might explore how to evaluate unlearning methods more comprehensively, such as whether the model could recall forgotten content or whether methods would make forgotten data identifiable.  In addition, we perform all the experiments in simulated settings. Future work might apply our methods to real-world applications to deal with actual use cases or introduce new benchmarks for evaluating unlearning methods.

\section*{Acknowledgment}
We would like to thank all reviewers and the SALT Lab for their valuable feedback.  
This work was partially sponsored by NSF grant IIS-2247357 and IIS-2308994.
\bibliography{anthology,custom}

\begin{thebibliography}{51}
\expandafter\ifx\csname natexlab\endcsname\relax\def\natexlab#1{#1}\fi

\bibitem[{Abadi et~al.(2016)Abadi, Chu, Goodfellow, McMahan, Mironov, Talwar,
  and Zhang}]{abadi2016deep}
Martin Abadi, Andy Chu, Ian Goodfellow, H~Brendan McMahan, Ilya Mironov, Kunal
  Talwar, and Li~Zhang. 2016.
\newblock Deep learning with differential privacy.
\newblock In \emph{Proceedings of the 2016 ACM SIGSAC conference on computer
  and communications security}, pages 308--318.

\bibitem[{Anil et~al.(2021)Anil, Ghazi, Gupta, Kumar, and
  Manurangsi}]{anil2021large}
Rohan Anil, Badih Ghazi, Vineet Gupta, Ravi Kumar, and Pasin Manurangsi. 2021.
\newblock Large-scale differentially private bert.
\newblock \emph{arXiv preprint arXiv:2108.01624}.

\bibitem[{Belinkov et~al.(2017)Belinkov, Durrani, Dalvi, Sajjad, and
  Glass}]{belinkov2017neural}
Yonatan Belinkov, Nadir Durrani, Fahim Dalvi, Hassan Sajjad, and James Glass.
  2017.
\newblock What do neural machine translation models learn about morphology?
\newblock \emph{arXiv preprint arXiv:1704.03471}.

\bibitem[{Bourtoule et~al.(2021)Bourtoule, Chandrasekaran, Choquette-Choo, Jia,
  Travers, Zhang, Lie, and Papernot}]{bourtoule2021machine}
Lucas Bourtoule, Varun Chandrasekaran, Christopher~A Choquette-Choo, Hengrui
  Jia, Adelin Travers, Baiwu Zhang, David Lie, and Nicolas Papernot. 2021.
\newblock Machine unlearning.
\newblock In \emph{2021 IEEE Symposium on Security and Privacy (SP)}, pages
  141--159. IEEE.

\bibitem[{Brown et~al.(2020)Brown, Mann, Ryder, Subbiah, Kaplan, Dhariwal,
  Neelakantan, Shyam, Sastry, Askell, Agarwal, Herbert{-}Voss, Krueger,
  Henighan, Child, Ramesh, Ziegler, Wu, Winter, Hesse, Chen, Sigler, Litwin,
  Gray, Chess, Clark, Berner, McCandlish, Radford, Sutskever, and
  Amodei}]{brown2020language}
Tom~B. Brown, Benjamin Mann, Nick Ryder, Melanie Subbiah, Jared Kaplan,
  Prafulla Dhariwal, Arvind Neelakantan, Pranav Shyam, Girish Sastry, Amanda
  Askell, Sandhini Agarwal, Ariel Herbert{-}Voss, Gretchen Krueger, Tom
  Henighan, Rewon Child, Aditya Ramesh, Daniel~M. Ziegler, Jeffrey Wu, Clemens
  Winter, Christopher Hesse, Mark Chen, Eric Sigler, Mateusz Litwin, Scott
  Gray, Benjamin Chess, Jack Clark, Christopher Berner, Sam McCandlish, Alec
  Radford, Ilya Sutskever, and Dario Amodei. 2020.
\newblock \href
  {https://proceedings.neurips.cc/paper/2020/hash/1457c0d6bfcb4967418bfb8ac142f64a-Abstract.html}
  {Language models are few-shot learners}.
\newblock In \emph{Advances in Neural Information Processing Systems 33: Annual
  Conference on Neural Information Processing Systems 2020, NeurIPS 2020,
  December 6-12, 2020, virtual}.

\bibitem[{Carlini et~al.(2022)Carlini, Ippolito, Jagielski, Lee, Tramer, and
  Zhang}]{carlini2022quantifying}
Nicholas Carlini, Daphne Ippolito, Matthew Jagielski, Katherine Lee, Florian
  Tramer, and Chiyuan Zhang. 2022.
\newblock Quantifying memorization across neural language models.
\newblock \emph{arXiv preprint arXiv:2202.07646}.

\bibitem[{Carlini et~al.(2021)Carlini, Tramer, Wallace, Jagielski,
  Herbert-Voss, Lee, Roberts, Brown, Song, Erlingsson
  et~al.}]{carlini2021extracting}
Nicholas Carlini, Florian Tramer, Eric Wallace, Matthew Jagielski, Ariel
  Herbert-Voss, Katherine Lee, Adam Roberts, Tom~B Brown, Dawn Song, Ulfar
  Erlingsson, et~al. 2021.
\newblock Extracting training data from large language models.
\newblock In \emph{USENIX Security Symposium}, volume~6.

\bibitem[{Chen et~al.(2022)Chen, Dodda, and Yang}]{chen2022humanintheloop}
Jiaao Chen, Mohan Dodda, and Diyi Yang. 2022.
\newblock \href {http://arxiv.org/abs/2212.09750} {Human-in-the-loop
  abstractive dialogue summarization}.

\bibitem[{Chen et~al.(2023)Chen, Zhang, Shi, Li, Smola, and
  Yang}]{chen2023parameter}
Jiaao Chen, Aston Zhang, Xingjian Shi, Mu~Li, Alex Smola, and Diyi Yang. 2023.
\newblock Parameter-efficient fine-tuning design spaces.
\newblock \emph{arXiv preprint arXiv:2301.01821}.

\bibitem[{Chien et~al.(2023)Chien, Pan, and Milenkovic}]{chien2023efficient}
Eli Chien, Chao Pan, and Olgica Milenkovic. 2023.
\newblock Efficient model updates for approximate unlearning of
  graph-structured data.
\newblock In \emph{The Eleventh International Conference on Learning
  Representations}.

\bibitem[{Chowdhery et~al.(2022{\natexlab{a}})Chowdhery, Narang, Devlin, Bosma,
  Mishra, Roberts, Barham, Chung, Sutton, Gehrmann, Schuh, Shi, Tsvyashchenko,
  Maynez, Rao, Barnes, Tay, Shazeer, Prabhakaran, Reif, Du, Hutchinson, Pope,
  Bradbury, Austin, Isard, Gur-Ari, Yin, Duke, Levskaya, Ghemawat, Dev,
  Michalewski, Garcia, Misra, Robinson, Fedus, Zhou, Ippolito, Luan, Lim, Zoph,
  Spiridonov, Sepassi, Dohan, Agrawal, Omernick, Dai, Pillai, Pellat,
  Lewkowycz, Moreira, Child, Polozov, Lee, Zhou, Wang, Saeta, Diaz, Firat,
  Catasta, Wei, Meier-Hellstern, Eck, Dean, Petrov, and Fiedel}]{palm}
Aakanksha Chowdhery, Sharan Narang, Jacob Devlin, Maarten Bosma, Gaurav Mishra,
  Adam Roberts, Paul Barham, Hyung~Won Chung, Charles Sutton, Sebastian
  Gehrmann, Parker Schuh, Kensen Shi, Sasha Tsvyashchenko, Joshua Maynez,
  Abhishek Rao, Parker Barnes, Yi~Tay, Noam Shazeer, Vinodkumar Prabhakaran,
  Emily Reif, Nan Du, Ben Hutchinson, Reiner Pope, James Bradbury, Jacob
  Austin, Michael Isard, Guy Gur-Ari, Pengcheng Yin, Toju Duke, Anselm
  Levskaya, Sanjay Ghemawat, Sunipa Dev, Henryk Michalewski, Xavier Garcia,
  Vedant Misra, Kevin Robinson, Liam Fedus, Denny Zhou, Daphne Ippolito, David
  Luan, Hyeontaek Lim, Barret Zoph, Alexander Spiridonov, Ryan Sepassi, David
  Dohan, Shivani Agrawal, Mark Omernick, Andrew~M. Dai,
  Thanumalayan~Sankaranarayana Pillai, Marie Pellat, Aitor Lewkowycz, Erica
  Moreira, Rewon Child, Oleksandr Polozov, Katherine Lee, Zongwei Zhou, Xuezhi
  Wang, Brennan Saeta, Mark Diaz, Orhan Firat, Michele Catasta, Jason Wei,
  Kathy Meier-Hellstern, Douglas Eck, Jeff Dean, Slav Petrov, and Noah Fiedel.
  2022{\natexlab{a}}.
\newblock \href {https://arxiv.org/abs/2204.02311} {Palm: Scaling language
  modeling with pathways}.

\bibitem[{Chowdhery et~al.(2022{\natexlab{b}})Chowdhery, Narang, Devlin, Bosma,
  Mishra, Roberts, Barham, Chung, Sutton, Gehrmann et~al.}]{chowdhery2022palm}
Aakanksha Chowdhery, Sharan Narang, Jacob Devlin, Maarten Bosma, Gaurav Mishra,
  Adam Roberts, Paul Barham, Hyung~Won Chung, Charles Sutton, Sebastian
  Gehrmann, et~al. 2022{\natexlab{b}}.
\newblock Palm: Scaling language modeling with pathways.
\newblock \emph{arXiv preprint arXiv:2204.02311}.

\bibitem[{Dai et~al.(2021)Dai, Dong, Hao, Sui, Chang, and
  Wei}]{dai2021knowledge}
Damai Dai, Li~Dong, Yaru Hao, Zhifang Sui, Baobao Chang, and Furu Wei. 2021.
\newblock Knowledge neurons in pretrained transformers.
\newblock \emph{arXiv preprint arXiv:2104.08696}.

\bibitem[{Dwork(2008)}]{dwork2008differential}
Cynthia Dwork. 2008.
\newblock Differential privacy: A survey of results.
\newblock In \emph{Theory and Applications of Models of Computation: 5th
  International Conference, TAMC 2008, Xi’an, China, April 25-29, 2008.
  Proceedings 5}, pages 1--19. Springer.

\bibitem[{Gliwa et~al.(2019)Gliwa, Mochol, Biesek, and Wawer}]{gliwa2019samsum}
Bogdan Gliwa, Iwona Mochol, Maciej Biesek, and Aleksander Wawer. 2019.
\newblock Samsum corpus: A human-annotated dialogue dataset for abstractive
  summarization.
\newblock \emph{arXiv preprint arXiv:1911.12237}.

\bibitem[{Golatkar et~al.(2020)Golatkar, Achille, and
  Soatto}]{golatkar2020eternal}
Aditya Golatkar, Alessandro Achille, and Stefano Soatto. 2020.
\newblock Eternal sunshine of the spotless net: Selective forgetting in deep
  networks.
\newblock In \emph{Proceedings of the IEEE/CVF Conference on Computer Vision
  and Pattern Recognition}, pages 9304--9312.

\bibitem[{Hoffmann et~al.(2022)Hoffmann, Borgeaud, Mensch, Buchatskaya, Cai,
  Rutherford, Casas, Hendricks, Welbl, Clark et~al.}]{hoffmann2022training}
Jordan Hoffmann, Sebastian Borgeaud, Arthur Mensch, Elena Buchatskaya, Trevor
  Cai, Eliza Rutherford, Diego de~Las Casas, Lisa~Anne Hendricks, Johannes
  Welbl, Aidan Clark, et~al. 2022.
\newblock Training compute-optimal large language models.
\newblock \emph{arXiv preprint arXiv:2203.15556}.

\bibitem[{Houlsby et~al.(2019)Houlsby, Giurgiu, Jastrzebski, Morrone,
  De~Laroussilhe, Gesmundo, Attariyan, and Gelly}]{houlsby2019parameter}
Neil Houlsby, Andrei Giurgiu, Stanislaw Jastrzebski, Bruna Morrone, Quentin
  De~Laroussilhe, Andrea Gesmundo, Mona Attariyan, and Sylvain Gelly. 2019.
\newblock Parameter-efficient transfer learning for nlp.
\newblock In \emph{International Conference on Machine Learning}, pages
  2790--2799. PMLR.

\bibitem[{Izzo et~al.(2021)Izzo, Smart, Chaudhuri, and
  Zou}]{izzo2021approximate}
Zachary Izzo, Mary~Anne Smart, Kamalika Chaudhuri, and James Zou. 2021.
\newblock Approximate data deletion from machine learning models.
\newblock In \emph{International Conference on Artificial Intelligence and
  Statistics}, pages 2008--2016. PMLR.

\bibitem[{Jagielski et~al.(2022)Jagielski, Thakkar, Tramer, Ippolito, Lee,
  Carlini, Wallace, Song, Thakurta, Papernot et~al.}]{jagielski2022measuring}
Matthew Jagielski, Om~Thakkar, Florian Tramer, Daphne Ippolito, Katherine Lee,
  Nicholas Carlini, Eric Wallace, Shuang Song, Abhradeep Thakurta, Nicolas
  Papernot, et~al. 2022.
\newblock Measuring forgetting of memorized training examples.
\newblock \emph{arXiv preprint arXiv:2207.00099}.

\bibitem[{Jang et~al.(2022)Jang, Yoon, Yang, Cha, Lee, Logeswaran, and
  Seo}]{jang2022knowledge}
Joel Jang, Dongkeun Yoon, Sohee Yang, Sungmin Cha, Moontae Lee, Lajanugen
  Logeswaran, and Minjoon Seo. 2022.
\newblock Knowledge unlearning for mitigating privacy risks in language models.
\newblock \emph{arXiv preprint arXiv:2210.01504}.

\bibitem[{Ji et~al.(2023)Ji, Lee, Frieske, Yu, Su, Xu, Ishii, Bang, Madotto,
  and Fung}]{ji2023survey}
Ziwei Ji, Nayeon Lee, Rita Frieske, Tiezheng Yu, Dan Su, Yan Xu, Etsuko Ishii,
  Ye~Jin Bang, Andrea Madotto, and Pascale Fung. 2023.
\newblock Survey of hallucination in natural language generation.
\newblock \emph{ACM Computing Surveys}, 55(12):1--38.

\bibitem[{Kojima et~al.(2022)Kojima, Gu, Reid, Matsuo, and
  Iwasawa}]{kojima2022large}
Takeshi Kojima, Shixiang~Shane Gu, Machel Reid, Yutaka Matsuo, and Yusuke
  Iwasawa. 2022.
\newblock \href {https://arxiv.org/abs/2205.11916} {Large language models are
  zero-shot reasoners}.
\newblock In \emph{Thirty-sixth Conference on Neural Information Processing
  Systems (NeurIPS 2022)}.

\bibitem[{Kumar et~al.(2022)Kumar, Gangadharaiah, and Roth}]{kumar2022privacy}
Vinayshekhar~Bannihatti Kumar, Rashmi Gangadharaiah, and Dan Roth. 2022.
\newblock Privacy adhering machine un-learning in nlp.
\newblock \emph{arXiv preprint arXiv:2212.09573}.

\bibitem[{Kurmanji et~al.(2023)Kurmanji, Triantafillou, and
  Triantafillou}]{kurmanji2023towards}
Meghdad Kurmanji, Peter Triantafillou, and Eleni Triantafillou. 2023.
\newblock Towards unbounded machine unlearning.
\newblock \emph{arXiv preprint arXiv:2302.09880}.

\bibitem[{Lee et~al.(2021)Lee, Ippolito, Nystrom, Zhang, Eck, Callison-Burch,
  and Carlini}]{lee2021deduplicating}
Katherine Lee, Daphne Ippolito, Andrew Nystrom, Chiyuan Zhang, Douglas Eck,
  Chris Callison-Burch, and Nicholas Carlini. 2021.
\newblock Deduplicating training data makes language models better.
\newblock \emph{arXiv preprint arXiv:2107.06499}.

\bibitem[{Lewkowycz et~al.(2022)Lewkowycz, Andreassen, Dohan, Dyer,
  Michalewski, Ramasesh, Slone, Anil, Schlag, Gutman-Solo
  et~al.}]{lewkowycz2022solving}
Aitor Lewkowycz, Anders Andreassen, David Dohan, Ethan Dyer, Henryk
  Michalewski, Vinay Ramasesh, Ambrose Slone, Cem Anil, Imanol Schlag, Theo
  Gutman-Solo, et~al. 2022.
\newblock Solving quantitative reasoning problems with language models.
\newblock \emph{arXiv preprint arXiv:2206.14858}.

\bibitem[{Li et~al.(2021)Li, Tramer, Liang, and Hashimoto}]{li2021large}
Xuechen Li, Florian Tramer, Percy Liang, and Tatsunori Hashimoto. 2021.
\newblock Large language models can be strong differentially private learners.
\newblock \emph{arXiv preprint arXiv:2110.05679}.

\bibitem[{Liang et~al.(2022)Liang, Bommasani, Lee, Tsipras, Soylu, Yasunaga,
  Zhang, Narayanan, Wu, Kumar et~al.}]{liang2022holistic}
Percy Liang, Rishi Bommasani, Tony Lee, Dimitris Tsipras, Dilara Soylu,
  Michihiro Yasunaga, Yian Zhang, Deepak Narayanan, Yuhuai Wu, Ananya Kumar,
  et~al. 2022.
\newblock \href {https://arxiv.org/abs/2211.09110} {Holistic evaluation of
  language models}.
\newblock \emph{ArXiv preprint}, abs/2211.09110.

\bibitem[{Liu et~al.(2022)Liu, Liu, and Stone}]{liu2022continual}
Bo~Liu, Qiang Liu, and Peter Stone. 2022.
\newblock Continual learning and private unlearning.
\newblock In \emph{Conference on Lifelong Learning Agents}, pages 243--254.
  PMLR.

\bibitem[{Maas et~al.(2011)Maas, Daly, Pham, Huang, Ng, and
  Potts}]{maas2011learning}
Andrew Maas, Raymond~E Daly, Peter~T Pham, Dan Huang, Andrew~Y Ng, and
  Christopher Potts. 2011.
\newblock Learning word vectors for sentiment analysis.
\newblock In \emph{Proceedings of the 49th annual meeting of the association
  for computational linguistics: Human language technologies}, pages 142--150.

\bibitem[{Min et~al.(2022)Min, Lyu, Holtzman, Artetxe, Lewis, Hajishirzi, and
  Zettlemoyer}]{min2022rethinking}
Sewon Min, Xinxi Lyu, Ari Holtzman, Mikel Artetxe, Mike Lewis, Hannaneh
  Hajishirzi, and Luke Zettlemoyer. 2022.
\newblock \href {https://arxiv.org/abs/2202.12837} {Rethinking the role of
  demonstrations: What makes in-context learning work?}
\newblock \emph{arXiv preprint arXiv:2202.12837}.

\bibitem[{Mitchell et~al.(2021)Mitchell, Lin, Bosselut, Finn, and
  Manning}]{mitchell2021fast}
Eric Mitchell, Charles Lin, Antoine Bosselut, Chelsea Finn, and Christopher~D
  Manning. 2021.
\newblock Fast model editing at scale.
\newblock \emph{arXiv preprint arXiv:2110.11309}.

\bibitem[{Nguyen et~al.(2022)Nguyen, Huynh, Nguyen, Liew, Yin, and
  Nguyen}]{nguyen2022survey}
Thanh~Tam Nguyen, Thanh~Trung Huynh, Phi~Le Nguyen, Alan Wee-Chung Liew,
  Hongzhi Yin, and Quoc Viet~Hung Nguyen. 2022.
\newblock A survey of machine unlearning.
\newblock \emph{arXiv preprint arXiv:2209.02299}.

\bibitem[{Ouyang et~al.(2022)Ouyang, Wu, Jiang, Almeida, Wainwright, Mishkin,
  Zhang, Agarwal, Slama, Ray et~al.}]{ouyang2022training}
Long Ouyang, Jeff Wu, Xu~Jiang, Diogo Almeida, Carroll~L Wainwright, Pamela
  Mishkin, Chong Zhang, Sandhini Agarwal, Katarina Slama, Alex Ray, et~al.
  2022.
\newblock Training language models to follow instructions with human feedback.
\newblock \emph{arXiv preprint arXiv:2203.02155}.

\bibitem[{Qin et~al.(2023)Qin, Zhang, Zhang, Chen, Yasunaga, and
  Yang}]{qin2023chatgpt}
Chengwei Qin, Aston Zhang, Zhuosheng Zhang, Jiaao Chen, Michihiro Yasunaga, and
  Diyi Yang. 2023.
\newblock Is chatgpt a general-purpose natural language processing task solver?
\newblock \emph{arXiv preprint arXiv:2302.06476}.

\bibitem[{Radford et~al.(2019)Radford, Wu, Child, Luan, Amodei, Sutskever
  et~al.}]{Radford2019LanguageMA}
Alec Radford, Jeffrey Wu, Rewon Child, David Luan, Dario Amodei, Ilya
  Sutskever, et~al. 2019.
\newblock Language models are unsupervised multitask learners.
\newblock \emph{OpenAI blog}, page~9.

\bibitem[{Rae et~al.(2021)Rae, Borgeaud, Cai, Millican, Hoffmann, Song,
  Aslanides, Henderson, Ring, Young et~al.}]{rae2021scaling}
Jack~W Rae, Sebastian Borgeaud, Trevor Cai, Katie Millican, Jordan Hoffmann,
  Francis Song, John Aslanides, Sarah Henderson, Roman Ring, Susannah Young,
  et~al. 2021.
\newblock Scaling language models: Methods, analysis \& insights from training
  gopher.
\newblock \emph{arXiv preprint arXiv:2112.11446}.

\bibitem[{Raffel et~al.(2020)Raffel, Shazeer, Roberts, Lee, Narang, Matena,
  Zhou, Li, and Liu}]{t5}
Colin Raffel, Noam Shazeer, Adam Roberts, Katherine Lee, Sharan Narang, Michael
  Matena, Yanqi Zhou, Wei Li, and Peter~J. Liu. 2020.
\newblock Exploring the limits of transfer learning with a unified text-to-text
  transformer.
\newblock \emph{JMLR}, 21(140):1--67.

\bibitem[{Sanh et~al.(2021)Sanh, Webson, Raffel, Bach, Sutawika, Alyafeai,
  Chaffin, Stiegler, Scao, Raja et~al.}]{sanh2021multitask}
Victor Sanh, Albert Webson, Colin Raffel, Stephen~H Bach, Lintang Sutawika,
  Zaid Alyafeai, Antoine Chaffin, Arnaud Stiegler, Teven~Le Scao, Arun Raja,
  et~al. 2021.
\newblock Multitask prompted training enables zero-shot task generalization.
\newblock \emph{arXiv preprint arXiv:2110.08207}.

\bibitem[{Sekhari et~al.(2021)Sekhari, Acharya, Kamath, and
  Suresh}]{sekhari2021remember}
Ayush Sekhari, Jayadev Acharya, Gautam Kamath, and Ananda~Theertha Suresh.
  2021.
\newblock Remember what you want to forget: Algorithms for machine unlearning.
\newblock \emph{Advances in Neural Information Processing Systems},
  34:18075--18086.

\bibitem[{Smith et~al.(2022)Smith, Patwary, Norick, LeGresley, Rajbhandari,
  Casper, Liu, Prabhumoye, Zerveas, Korthikanti et~al.}]{smith2022using}
Shaden Smith, Mostofa Patwary, Brandon Norick, Patrick LeGresley, Samyam
  Rajbhandari, Jared Casper, Zhun Liu, Shrimai Prabhumoye, George Zerveas,
  Vijay Korthikanti, et~al. 2022.
\newblock Using deepspeed and megatron to train megatron-turing nlg 530b, a
  large-scale generative language model.
\newblock \emph{arXiv preprint arXiv:2201.11990}.

\bibitem[{Touvron et~al.(2023)Touvron, Lavril, Izacard, Martinet, Lachaux,
  Lacroix, Rozi{\`e}re, Goyal, Hambro, Azhar et~al.}]{touvron2023llama}
Hugo Touvron, Thibaut Lavril, Gautier Izacard, Xavier Martinet, Marie-Anne
  Lachaux, Timoth{\'e}e Lacroix, Baptiste Rozi{\`e}re, Naman Goyal, Eric
  Hambro, Faisal Azhar, et~al. 2023.
\newblock Llama: Open and efficient foundation language models.
\newblock \emph{arXiv preprint arXiv:2302.13971}.

\bibitem[{Wang et~al.(2020)Wang, Tang, Duan, Wei, Huang, Cao, Jiang, Zhou
  et~al.}]{wang2020k}
Ruize Wang, Duyu Tang, Nan Duan, Zhongyu Wei, Xuanjing Huang, Guihong Cao,
  Daxin Jiang, Ming Zhou, et~al. 2020.
\newblock K-adapter: Infusing knowledge into pre-trained models with adapters.
\newblock \emph{arXiv preprint arXiv:2002.01808}.

\bibitem[{Wang et~al.(2021)Wang, Fang, Khabsa, Mao, and
  Ma}]{wang2021entailment}
Sinong Wang, Han Fang, Madian Khabsa, Hanzi Mao, and Hao Ma. 2021.
\newblock Entailment as few-shot learner.
\newblock \emph{arXiv preprint arXiv:2104.14690}.

\bibitem[{Wang et~al.(2022)Wang, Kordi, Mishra, Liu, Smith, Khashabi, and
  Hajishirzi}]{wang2022self}
Yizhong Wang, Yeganeh Kordi, Swaroop Mishra, Alisa Liu, Noah~A Smith, Daniel
  Khashabi, and Hannaneh Hajishirzi. 2022.
\newblock Self-instruct: Aligning language model with self generated
  instructions.
\newblock \emph{arXiv preprint arXiv:2212.10560}.

\bibitem[{Webson and Pavlick(2022)}]{webson2021prompt}
Albert Webson and Ellie Pavlick. 2022.
\newblock \href {https://doi.org/10.18653/v1/2022.naacl-main.167} {Do
  prompt-based models really understand the meaning of their prompts?}
\newblock In \emph{Proceedings of the 2022 Conference of the North American
  Chapter of the Association for Computational Linguistics: Human Language
  Technologies}, pages 2300--2344, Seattle, United States. Association for
  Computational Linguistics.

\bibitem[{Wei et~al.(2021)Wei, Bosma, Zhao, Guu, Yu, Lester, Du, Dai, and
  Le}]{wei2021finetuned}
Jason Wei, Maarten Bosma, Vincent~Y Zhao, Kelvin Guu, Adams~Wei Yu, Brian
  Lester, Nan Du, Andrew~M Dai, and Quoc~V Le. 2021.
\newblock Finetuned language models are zero-shot learners.
\newblock \emph{arXiv preprint arXiv:2109.01652}.

\bibitem[{Yang et~al.(2019)Yang, Dai, Yang, Carbonell, Salakhutdinov, and
  Le}]{yang2019xlnet}
Zhilin Yang, Zihang Dai, Yiming Yang, Jaime Carbonell, Russ~R Salakhutdinov,
  and Quoc~V Le. 2019.
\newblock Xlnet: Generalized autoregressive pretraining for language
  understanding.
\newblock \emph{Advances in neural information processing systems}, 32.

\bibitem[{Yu et~al.(2021)Yu, Naik, Backurs, Gopi, Inan, Kamath, Kulkarni, Lee,
  Manoel, Wutschitz et~al.}]{yu2021differentially}
Da~Yu, Saurabh Naik, Arturs Backurs, Sivakanth Gopi, Huseyin~A Inan, Gautam
  Kamath, Janardhan Kulkarni, Yin~Tat Lee, Andre Manoel, Lukas Wutschitz,
  et~al. 2021.
\newblock Differentially private fine-tuning of language models.
\newblock \emph{arXiv preprint arXiv:2110.06500}.

\bibitem[{Zhou et~al.(2023)Zhou, Liu, Xu, Iyer, Sun, Mao, Ma, Efrat, Yu, Yu
  et~al.}]{zhou2023lima}
Chunting Zhou, Pengfei Liu, Puxin Xu, Srini Iyer, Jiao Sun, Yuning Mao, Xuezhe
  Ma, Avia Efrat, Ping Yu, Lili Yu, et~al. 2023.
\newblock Lima: Less is more for alignment.
\newblock \emph{arXiv preprint arXiv:2305.11206}.

\end{thebibliography}
\bibliographystyle{acl_natbib}



\end{document}